\documentclass[10pt]{article}

\usepackage{newtxtext,newtxmath}
\usepackage{booktabs}
\usepackage{siunitx}        
\usepackage{longtable}
\usepackage{threeparttable}
\usepackage{caption}
\usepackage{pgfplotstable}  
\pgfplotsset{compat=1.18}
\usepackage{graphicx,svg}
\usepackage{subcaption} 
\usepackage{booktabs}
\usepackage{subcaption} 
\usepackage{caption}
\usepackage{times}
\usepackage{url}
\usepackage[hidelinks]{hyperref}
\usepackage{booktabs,longtable,array,tabularx,threeparttable}

\usepackage[letterpaper,margin=1in]{geometry}
\usepackage{multirow}
\usepackage{array}
\usepackage{tikz}
\usetikzlibrary{calc}
\usetikzlibrary{shapes}

\usepackage{amssymb} 
\usepackage{setspace}
\usepackage[font=small,labelfont=bf]{caption}

\DeclareCaptionFont{singlespacing}{\setstretch{1}}  
\renewenvironment{abstract}
	{\quotation}
{\endquotation}

\date{}

\makeatletter
\renewcommand{\fnum@figure}{\textbf{Figure \thefigure}}
\renewcommand{\fnum@table}{\textbf{Table \thetable}}
\makeatother

\usepackage{scicite}
\usepackage{url}

\usepackage{xurl}

\renewcommand{\thetable}{\arabic{table}}
\renewcommand{\thefigure}{\arabic{figure}}

\def\scititle{Generative AI floods and dilutes the market for books}
\title{\bfseries \boldmath \scititle}

\author{
	Tuhin Chakrabarty$^{1}$, 
    Xinyue Liu $^{1}$, 
	Jane C. Ginsburg$^{2}$,
	Paramveer Dhillon$^{3,4}$\and
	\small$^{1}$Department of Computer Science, Stony Brook University.\\
        \small$^{2}$Columbia Law School.\\
	\small$^{3}$School of Information Science, University of Michigan.\\
    \small$^{4}$MIT Initiative on the Digital Economy.\and
	\small Corresponding authors:\\ \small tchakrabarty@cs.stonybrook.edu, \small ginsburg@law.columbia.edu, dhillonp@umich.edu
}

\begin{document} 

\maketitle

\begin{abstract} \bfseries \boldmath
Generative AI can produce book-length works of fiction at near-zero cost. These books are often dismissed as low-quality ``slop'' that buyers will ignore, and are assumed to carry little commercial weight. We test that assumption with full-text AI detection across 14,419 self-published genre-fiction books sold on Amazon from 2023 to 2026, matched to daily sales records through June 2026. None of these books disclose whether or not they contain AI-produced content. We find that books for which we detected substantial AI text ($>$ 25\%) make up a large share of the catalog but a smaller share of sales. Even so, they reach commercial scale, winning a growing share of sales over time and taking more of the scarce top-rank positions once held by books with no detected AI text. Over this period, the number of books with observed sales in a quarter grew 19.2-fold, while quarterly revenue grew only 8.9-fold. The market therefore added selling books faster than it added revenue, and revenue per selling book fell across most genres. Books with no AI text lose the most ground in genres with high AI diffusion, and most of all where Kindle Unlimited availability is high. Among top-selling books, those with substantial AI text draw on more distinctive language from existing books than do books with no AI text; for these books overlap rises with revenue, a gradient we do not detect for books with no AI text. Generative AI can thus reshape a creative market through scale rather than quality. Our results bear directly on the market-effect question at the center of the fair use defense to copyright infringement.
\end{abstract}
\noindent

\section{Introduction}

Generative AI has significantly reduced the cost of producing long form, book-length fiction \cite{pham_chang_iyyer_2026}. Online communities have formed around it \cite{gupta2026ai}, where people trade detailed recipes for turning a one-line premise into a finished novel.\footnote{See, for example, the communities at \url{https://www.reddit.com/r/WritingWithAI/} and \url{https://plotprose.com/}.} Readers and critics, however, often dismiss the output as ``slop” \cite{shaib2025measuring}, referring to it as cheap and formulaic work that buyers will ignore, thereby treating those books as commercially irrelevant. However, a book being low average quality and it being irrelevant are not necessarily the same thing. Due to its inexpensive nature, hundreds of producers can draft and list a book that is partially or fully generated by AI. And at that volume even an average product can attract the attention of a buyer and siphon away the sales that would otherwise go elsewhere. It is also worth acknowledging that many of these books reach readers without disclosure \cite{russell2026ai} and the brief previews shown on platforms make it hard for readers to discern which books are AI generated. It is, then, an empirical question whether low-quality, cheap books can still reshape a market. In 2026, Hachette canceled the horror novel \textit{Shy Girl} after an independent analysis found it to be largely AI-generated \cite{alter2026shygirl1, alter2026shygirl2}. One author, identified as ``Coral Hart,'' reportedly released more than 200 romance novels across 21 pen names in a single year and sold roughly 50{,}000 copies \cite{alter2026fabio}. These episodes provide strong evidence that such incidents can happen in real life, but individual cases cannot account for the market-wide effects. This paper is primarily aimed at measuring how generative AI has impacted a real book market once its output reached the catalog and competed for sales.

We study self-published genre fiction that is primarily sold on Amazon.\footnote{A small number of books that are traditionally published by other houses (roughly 0.3\% of the corpus) were identified during publisher verification. We decided to keep them as such in our analysis.} Self-publishing now makes up the bulk of books produced in the U.S. The number of self-published titles in the U.S.\ registered with Bowker crossed 3.5~million in 2025 (a 38.7\% rise over 2024) and accounted for roughly 84\% of the 4.17~million ISBN-registered titles released that year~\cite{milliot2026fourmillion}. Most of these books are released on the market through Amazon's Kindle Direct Publishing (KDP) platform. Unlike traditional publishing, KDP allows producers to add titles at a low marginal cost because there is no gatekeeping involved and the listing is nearly free~\cite{wordsrated2023amazon}. Within this broader market, genre fiction faces a high risk of being disrupted by generative AI. Readers of genre fiction come back for the same tropes and an influx of new books, which can easily be churned out by generative AI \cite{russell2026storyscope,chakrabarty2024art}. Given this, we expect self-published genre fiction to be one of the first creative markets where AI-written text appears at scale. To study this further in depth, we collect a dataset of 14,419 randomly selected self-published ebooks belonging to genre fiction that were published between January 2023 and March 2026. In addition to the books, we also collect their daily sales through June 2026.\footnote{A few of these titles have since been picked up by traditional publishing houses over the study period.} For each title we gather daily sales from a proprietary panel maintained by a major ``Big Five'' publisher. This panel tracks unit sales, revenue, price, and sales rank for approximately 500,000 ASINs on Amazon that make up approximately 95\% or more of daily unit volume in the eBook market. As a next step, we classified each book by the proportion of text in it that is detected as not human written, and clustered titles into those with no AI text, light AI text ($\leq 25\%$) and substantial AI text ($>25\%$).\footnote{We secured full texts for each of the selected books in our corpus lawfully by emailing authors for epubs, borrowing available ebooks online from local libraries such as Libby/Hoopla, as well as purchasing remaining ebooks from Amazon.} Finally, we matched each ebook to its Kindle Unlimited status, genre, and author metadata.

In a recent study, Reimers and Waldfogel \cite{reimers2026ai} showed how the number of new book releases has almost tripled after 2022 and that books that are detected as AI (based on 1000 word previews) earn lower ratings and weaker sales compared to books that are fully human written. In contrast, we are interested in the question of what happens to the market that receives these books. While they detect AI from previews and study how flagged titles themselves perform, we detect AI across the full text of each book and measure what those titles do to the books around them. When a massive number of books that are cheap to produce enter the market at a rate faster than the sales, revenue and top ranks available to absorb them, the return and attention left for the average book falls. We call this dilution. The same dynamic is now a central question in copyright litigation over AI training, where market effect has become a key, perhaps decisive, issue. In Kadrey v. Meta \cite{kadrey2025meta}, Judge Chhabria singled out market dilution, the erosion of a market through rapid production of substitutable works, as a theory that could determine the outcome of a fair-use defense in such cases, even as he noted that the supporting evidence was absent in the case before him \cite{kadrey2025meta}.

We therefore ask four questions. First, do books with substantial AI text only occupy the bottom of the market, or are they competitive enough that they sell and rank alongside books with no AI text in them? Second, do they expand the market or dilute it, by taking away sales and rank from books with no AI text? Third, do producers who use AI release books at inordinately high rates beyond what human-only writing would enable? Finally, among the successful books in terms of revenue, do those with substantial AI text draw on more distinctive language of existing books than those with no AI text? Our findings show a market rebuilt by scale. Books containing substantial AI text enter the catalog faster than the revenue available to absorb them and in turn decrease the revenue earned by individual books while competing for the top ranks once held by books with no AI text. Last but not least, books with AI text that succeed draw most heavily on the distinctive language of existing books. This restructuring of a creative labor market bears directly on the market-effect question now central to copyright law.
\section{Results}

We classify each book by the share of its full text detected as AI. Books designated as ``no AI text" carry no detected AI text, while books classified as ``light AI text" have scores above 0 and at most 25\%, and books with ``substantial AI text" exceed 25\%. The market-level results we report are robust to varying this 25\% cutoff.

\subsection{Books with substantial AI text spread across every genre but earn a smaller share of sales}

\begin{figure*}[!htbp]
\centering
\includegraphics[width=\textwidth]{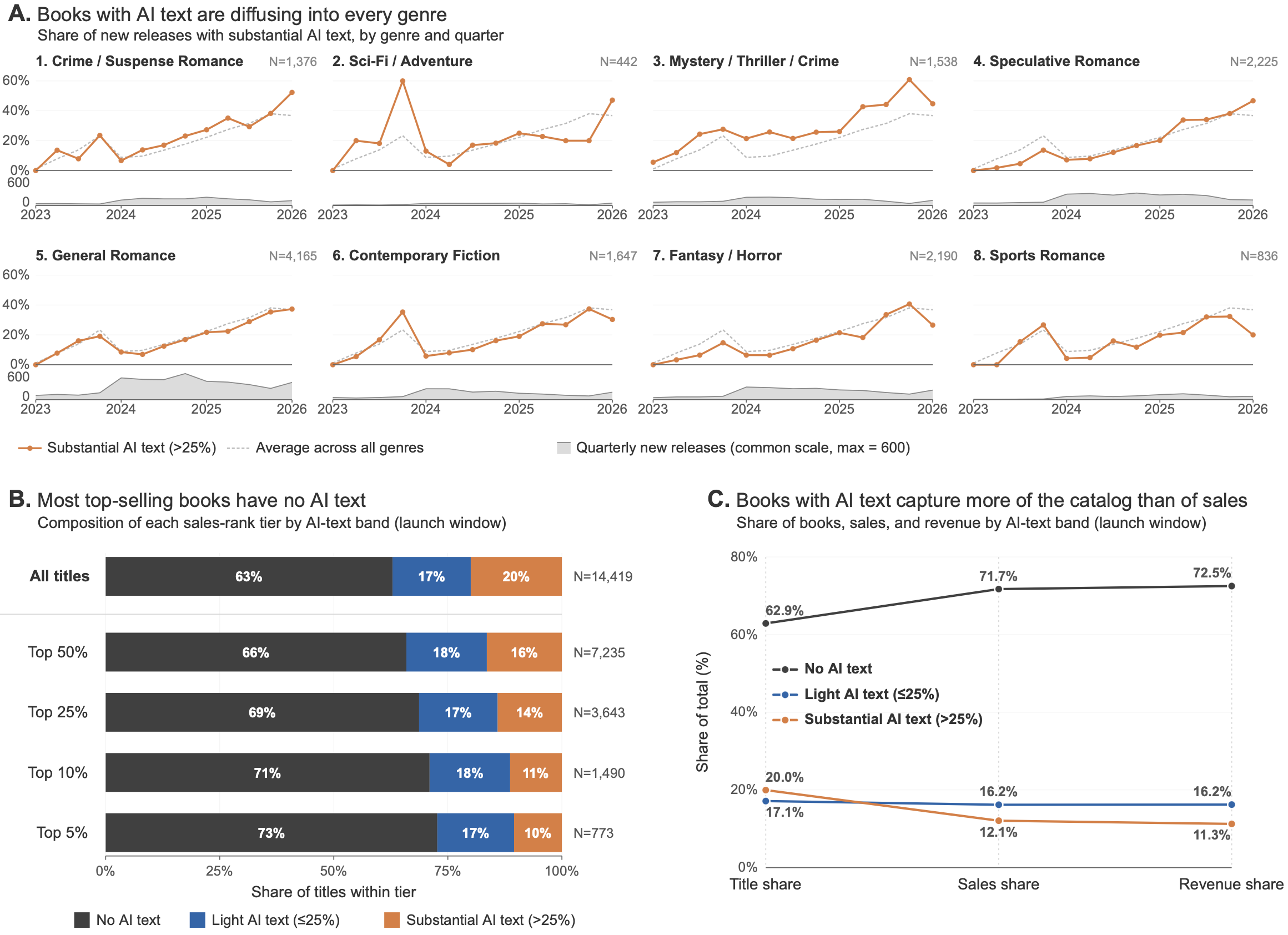}
\caption{\textbf{Books with substantial AI text spread across every genre yet earn a smaller share of sales.}
\textbf{(a)} Quarterly share of new releases with substantial AI text within each of the eight genre clusters, 2023--2026 (orange line, percentage scale), with the all-genre average shown for reference (dashed grey line). The shaded area gives the count of new releases in each genre in a given quarter. The share rises in every genre over the period. $N$ denotes the number of books in each genre cluster. \textbf{(b)} Composition of successive early-sales tiers, measured over the launch window. Tiers are formed by ranking books within genre-quarter cells and pooling, so tier counts differ slightly from exact percentages of $N$ (SI S4.2). Each bar gives the share of books with no AI text (dark), light AI text (blue), and substantial AI text (orange) within a tier, from all books ($N=14{,}419$) up to the top 5\% by early sales ($N=773$). The share of books with no AI text rises from 63\% of all books to 73\% of the top 5\%, while the share with substantial AI text falls from 20\% to 10\%. \textbf{(c)} Slopegraph showing the share of books, sales, and revenue held by each group over the launch window. Books with no AI text account for 62.9\% of books but a larger 71.7\% of sales and 72.5\% of revenue, whereas books with substantial AI text account for 20.0\% of books but only 12.1\% of sales and 11.3\% of revenue. The share of books with light AI text is near-constant across the three measures, at 17.1\% of books, 16.2\% of sales, and 16.2\% of revenue. Books are classified by the share of full text detected as AI: no AI text (0\%), light AI text (above 0 and at most 25\%), or substantial AI text (above 25\%).}
\label{fig:diffusion}
\end{figure*}

Books with substantial AI text have spread into every genre we track. Between January 2023 and March 2026, they made up a rising share of new releases across all eight genre clusters, \footnote{General Romance, Speculative / Adventure Romance, Crime / Suspense Romance, Sports Romance, Fantasy / Supernatural / Horror, Mystery / Thriller / Crime, Science Fiction / Adventure / Dystopian, and General / Contemporary Fiction.} and the all-genre average rose with them (Fig.~\ref{fig:diffusion}a). Their numbers did not translate into sales at the same rate. Books with substantial AI text make up 20\% of all books in our sample but earn only 12.1\% of sales and 11.3\% of revenue, while books with no AI text account for 62.9\% of books and a larger 71.7\% of sales and 72.5\% of revenue (Fig.~\ref{fig:diffusion}c). Books with light AI text fall between the two and account for 17.1\% of books and a near-equal 16.2\% of sales and 16.2\% of revenue. This implies that the shortfall in sales falls on the books with substantial AI text and does not depend on detected AI text in general. The gap becomes wider toward the top of the market. Books with no AI text hold 63\% of all launch-window\footnote{The launch window is a common observation window we use to standardize each book's sales. It covers a book's pre-release sales together with a fixed 90 days of sales after release, so that books released at different times are compared over the same horizon.} positions and 73\% of the best-selling top 5\%, while the share held by books with substantial AI text falls from 20\% of all books to 10\% of that top 5\% (Fig.~\ref{fig:diffusion}b). In other words, books with no AI text still hold most of the top positions. But it is worth noting that even a 12\% share of sales is crucial, especially when it comes from 20\% of the catalog that competes for a revenue pool which grew far more slowly than the catalog itself. In the next section we show how that share rose over the period and reached the top ranks.

\subsection{Books with substantial AI text reach commercial scale and the top ranks}

\begin{figure*}[htbp]
  \centering
  \includegraphics[width=\linewidth]{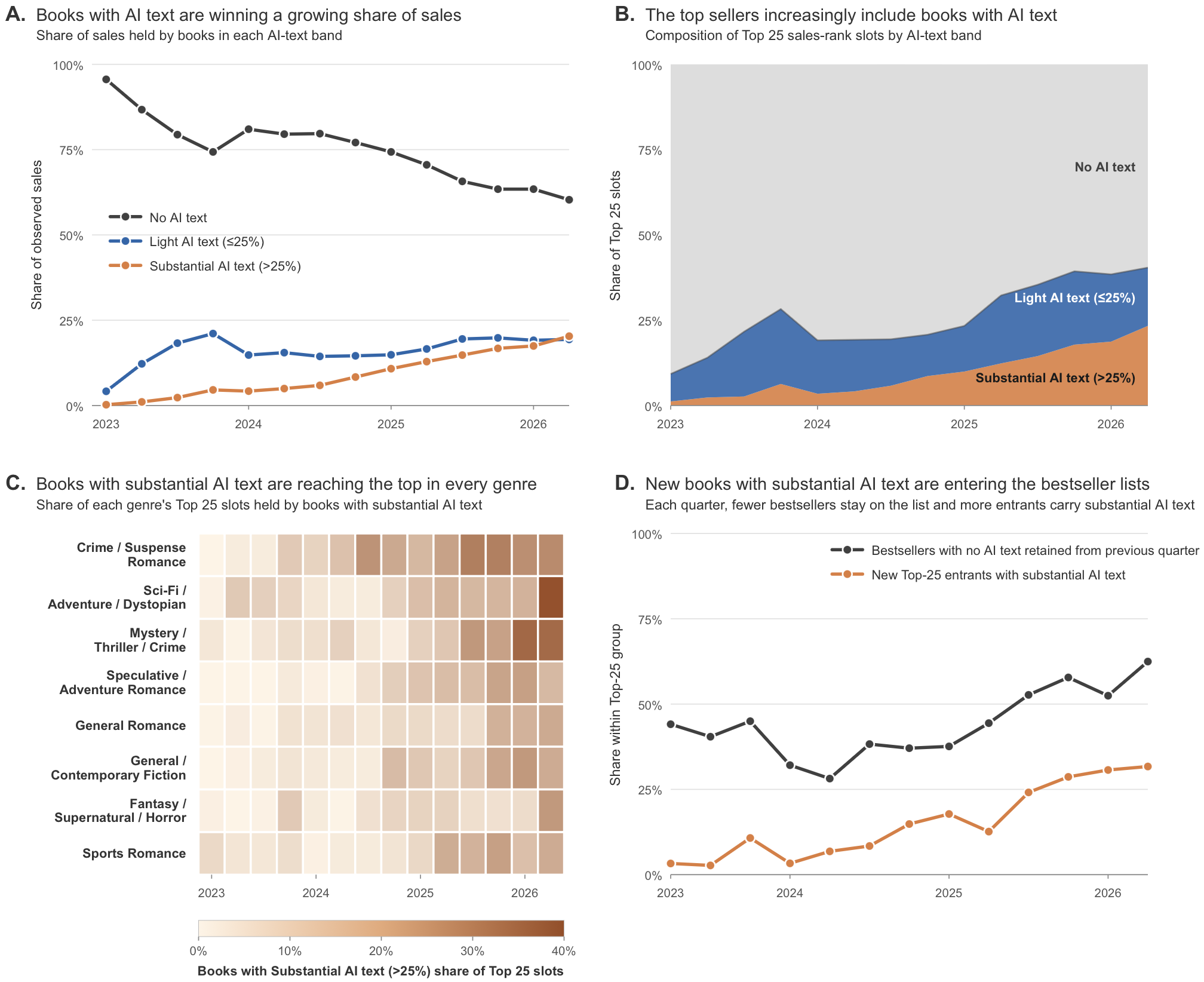}
 \caption{\textbf{Books with substantial AI text reach commercial scale and the top sales ranks.}
\textbf{(a)} Share of observed sales held by each AI text band, by quarter, 2023--2026. The share held by books with no AI text (black) declines from near 100\% in early 2023 to roughly 60\% by 2026 Q2, while the share with substantial AI text (orange) rises from near zero to about 20\% and the share with light AI text (blue) rises to about 19\%. \textbf{(b)} Composition of constructed Top-25 rank slots by AI text band. In each genre-week, sampled titles are ranked by median Amazon sales rank and the 25 highest-ranked sampled titles are retained; weekly slots are then aggregated to quarters. The combined share of books with light and substantial AI text grows from near zero in early 2023 to roughly 40\% by 2026 Q2. 
\textbf{(c)} Share of each genre cluster's constructed weekly Top-25 rank slots held by books with substantial AI text, aggregated to quarters. Darker cells denote a higher share. The share rises across the period in every cluster, and most steeply in Crime / Suspense Romance, Sci-Fi / Adventure / Dystopian, and Mystery / Thriller / Crime. 
\textbf{(d)} Turnover in constructed Top-25 rank sets. In each genre-quarter, sampled titles are ranked by median Amazon sales rank and the 25 highest-ranked sampled titles are retained. No-AI retention (black) falls to a trough near 28\% before rising to about 62\%. The share of new entrants with substantial AI text (orange) meanwhile rises from near zero to 31\%. The two lines measure different quantities and need not sum to 100\%.}
  \label{fig:reallocation}
\end{figure*}

Although such books sell weakly on average on their own, books with substantial AI text did not stay at the margins. Their share of observed sales rose over the period, from near zero in early 2023 to roughly 20\% by 2026 Q2, and the share held by books with light AI text rose alongside it (Fig.~\ref{fig:reallocation}a). These books also reached the scarce positions at the top of the market. Their share of constructed Top-25 rank slots grew throughout the period (Fig.~\ref{fig:reallocation}b).\footnote{For each genre and week, we rank titles in our sample by median Amazon sales rank and retain the 25 highest-ranked sampled titles, or all sampled titles if fewer than 25 have valid ranks. These weekly rank slots are then aggregated to quarters.} The same rise appears in every genre cluster (Fig.~\ref{fig:reallocation}c). As this happened, the top ranks turned over more quickly. After the first comparable quarter, retention of books with no AI text in the Top-25 fell to a trough near 28\% and recovered only unevenly thereafter, ending near 62\%, while the share of new Top-25 entrants with substantial AI text rose from near zero to 31\% (Fig.~\ref{fig:reallocation}d). Thus, turnover in the Top-25 sets increasingly brought in books with substantial AI text in them, even though the majority of the top-ranked sampled books remained the ones with no AI text in them. The market now holds many books with substantial AI text that sell little, along with a smaller set that reaches the very top.

\subsection{Releases outrun sales and revenue, and the decline reaches books with no AI text}

\begin{figure*}[htbp]
  \centering
  \includegraphics[width=\linewidth]{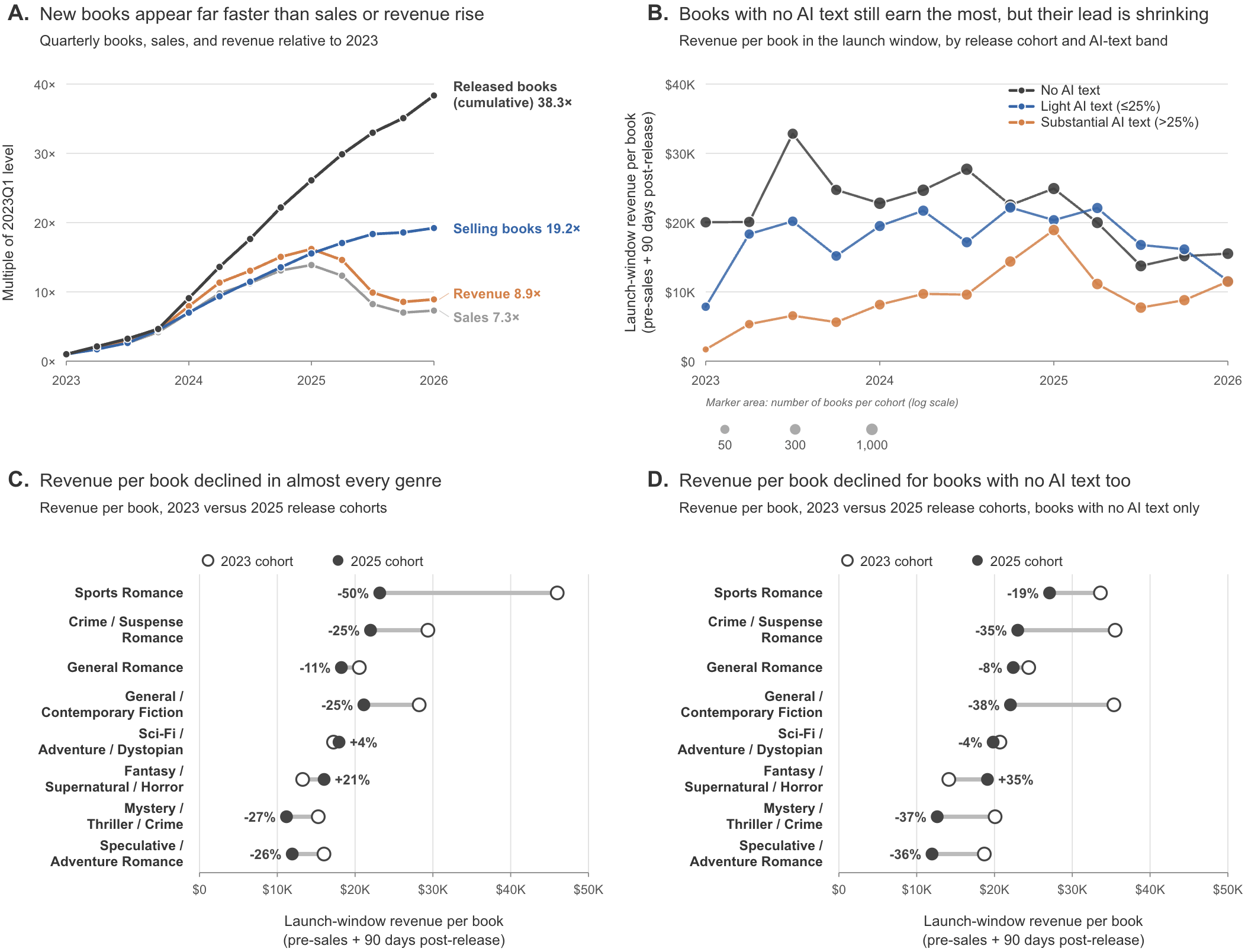}
\caption{\textbf{Releases outrun sales and revenue, and the decline reaches books with no AI text.}
\textbf{(a)} Catalog stock and quarterly market flows, each shown as a multiple of its 2023 Q1 level. The released-books line is cumulative, while selling titles, unit sales, and revenue are measured within each quarter. By 2026 Q1, the cumulative released catalog became 38.3 times its 2023 Q1 level. In an equivalent quarterly comparison, selling titles grew 19.2-fold, while unit sales and revenue grew only 7.3-fold and 8.9-fold. The number of books competing for sales therefore grew much faster than the market available to them. \textbf{(b)} Launch-window revenue per selling title, by release cohort and AI text band, with marker size scaled to the number of selling titles in each cohort (log scale). The launch window includes any pre-order of these books before the on-sale date and days 0 through 90 after it. The lead in revenue that books with no AI text (black) held over books with light AI text (blue) and substantial AI text (orange) narrows across later cohorts, as the returns to books with no AI text fall toward those of the other two bands. \textbf{(c)} Change in launch-window revenue per selling title between the 2023 and 2025 release cohorts, by genre cluster, for all books. Open circles mark the 2023 cohort and filled circles the 2025 cohort. Revenue per book fell in six of eight clusters, rising only in Fantasy / Supernatural / Horror (+21\%) and, marginally, Science Fiction / Adventure / Dystopian (+4\%). \textbf{(d)} The same launch-window comparison restricted to books with no AI text. Revenue per book fell in seven of eight clusters and rose only in Fantasy / Supernatural / Horror (+35\%). Because there is decline for books with no AI text alone, it does not arise from adding lower-revenue books with substantial AI text to the pool.}
\label{fig:dilution}
\end{figure*}

The number of books released expanded faster than the sales and revenue available to absorb them. Between January 2023 and March 2026, the cumulative catalog of released books grew 38.3-fold and the number of books selling in a quarter grew 19.2-fold. Meanwhile, the quarterly sales and revenue grew only 7.3-fold and 8.9-fold (Fig.~\ref{fig:dilution}a). Far more books now compete for revenue that grew at a fraction of that pace, so the revenue each selling book earns has fallen.

To ensure a fixed observation window despite the catalog expansion, we measure revenue over the launch window and group books into release cohorts. The launch window includes any pre-order activity before the on-sale date and days 0 to 90 after it. We compare the same window across cohorts, such that later cohorts are not penalized for being on sale for a shorter amount of time. Across these fixed launch windows, revenue per book that sold at least one copy declined on net from the 2023 cohort to the 2025 cohort, after a rise in 2024 that reached every band. Across cohorts, the revenue lead that books with no AI text held over books with substantial AI text narrowed, as returns to books with no AI text fell toward those of the other bands (Fig.~\ref{fig:dilution}b). The decline reaches across the market. Revenue per book fell in six of eight genres for all books (Fig.~\ref{fig:dilution}c) and in seven of eight for books with no AI text alone (Fig.~\ref{fig:dilution}d), when comparing the 2023 and 2025 launch cohorts. The drop in revenue per book cannot be explained as a mix shift where the average drops only because the catalog is filled with many lower-revenue AI books, because the books with no AI text even earn less in later cohorts as well.

\begin{figure*}[htbp]
  \centering
  \includegraphics[width=\linewidth]{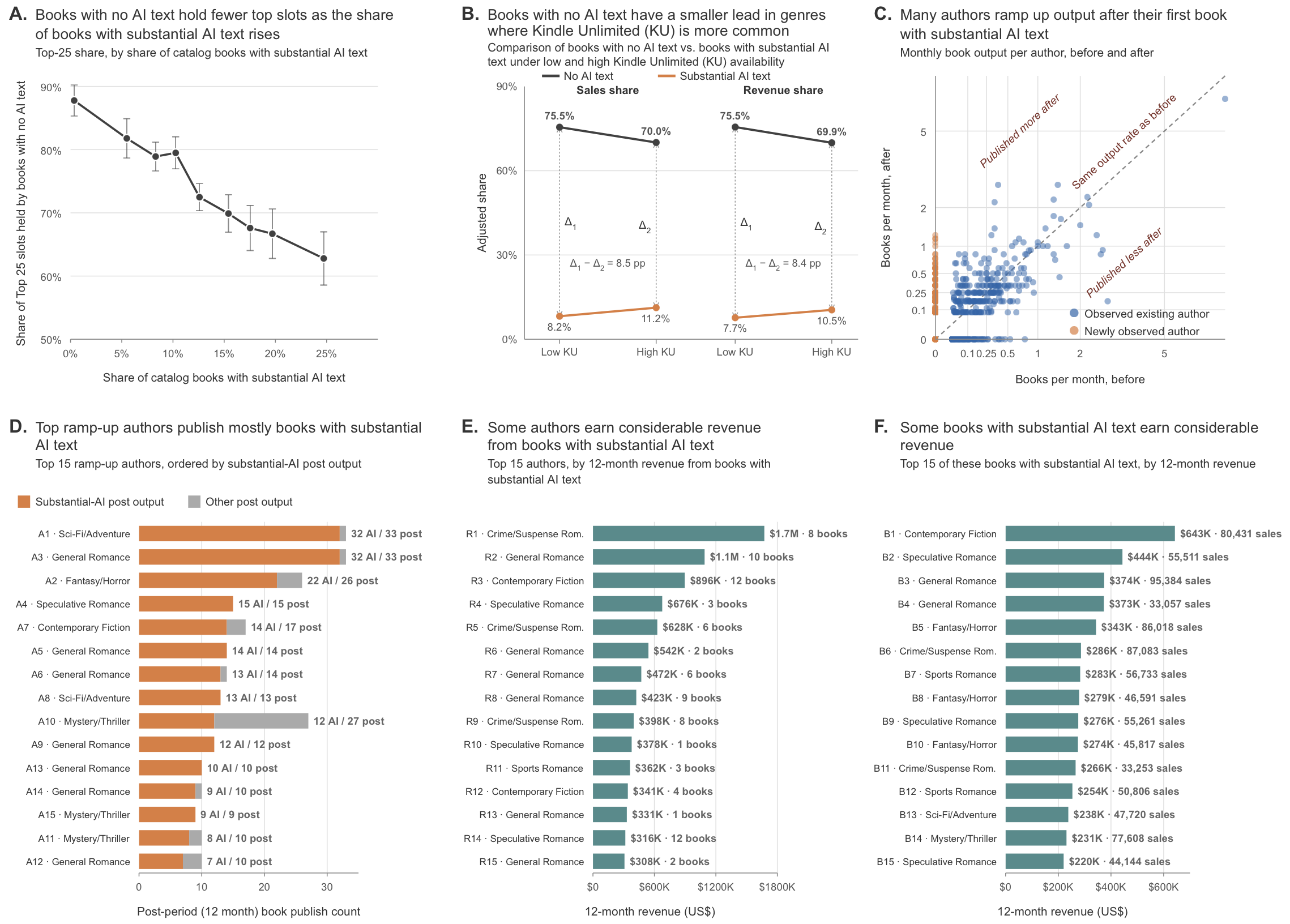}
\caption{\textbf{Losses for books with no AI text concentrate where AI exposure is greatest, and a broad base of authors scales output.}
\textbf{(a)} Share of Top 25 sales-rank slots held by books with no AI text against the share of the released catalog with substantial AI text, across genre-months grouped into exposure bins. The share held by books with no AI text falls from about 88\% in the lowest-exposure bin to about 63\% in the highest. Error bars show 95\% confidence intervals. \textbf{(b)} The lead that books with no AI text hold over books with substantial AI text in adjusted sales and revenue share, for genres grouped by Kindle Unlimited availability. The lead is 8.5 percentage points smaller in sales share, and 8.4 points smaller in revenue share, in genres where more titles are available on Kindle Unlimited than in genres where fewer are. \textbf{(c)} Monthly book output after each author's first substantial-AI text release against output over its observed pre-adoption period. The post-period covers months +1 through +12, with rates calculated over the post-adoption months observed for each author; month 0 is excluded.
Points above the diagonal mark bylines whose monthly output rose after that release. Blue points mark bylines with output before adoption; orange points on the vertical axis mark bylines first seen at adoption. Of the 824 authors in the event-time panel, 311 sit above the diagonal. Among the 385 authors who released further substantial-AI text books, 287 raised their output. \textbf{(d)} The 15 authors with the largest increases in monthly output after adoption, showing substantial-AI text books (orange) and other books (grey) published during the post-period. Most of these authors published almost entirely books with substantial AI text, with a few exceptions that mixed in. \textbf{(e)} The 15 author identities (or producers) whose substantial-AI text books generated the most gross consumer revenue during months +1 through +12 after each author's first such release. Books associated with the top author generated \$1.7M across eight titles, and books associated with eight of the 15 authors generated more than \$400K. Revenue is gross consumer spending before platform commissions and other deductions; author royalties and net earnings are lower. \textbf{(f)} The 15 individual books with substantial AI text earning the most revenue in the 12 months after release, with each book's revenue and unit sales. The top book earned \$643K on 80{,}431 sales, and each of the 15 earned at least \$220K.}
  \label{fig:exposure}
\end{figure*}

The losses become more prominent where exposure to AI is highest. Across genre-months, the share of Top-25 rank slots held by books with no AI text falls as the share of the released catalog with substantial AI text rises, from about 88\% in the lowest-exposure bin to about 63\% in the highest (Fig.~\ref{fig:exposure}a). The fall is steeper in genres where more of the titles are available on Kindle Unlimited. There, readers draw from a shared subscription pool, so each new book with substantial AI text competes for the same borrows. In these genres, the lead that books with no AI text hold over books with substantial AI text is 8.5 percentage points smaller in sales share, and 8.4 points smaller in revenue share, than in genres where fewer titles are available on Kindle Unlimited~(Fig.~\ref{fig:exposure}b). The same pattern holds at the low-exposure end.  In Fantasy / Supernatural / Horror, the genre AI reached latest and least, revenue per book for books with no AI text rose 35\% across cohorts even as it fell in the high-exposure genres. Returns to books with no AI text thus move with AI exposure across the market. It declines where exposure is high and holds steady where it is low. While observational, these patterns align with what dilution would predict.

\subsection{AI adoption increases author output across a broad base of producers}

The rise in output can be attributed to adopters who stuck with AI, instead of every author who tried it.  Within that group, output rose for a wide set of authors rather than a few high-volume producers, though the largest producers accounted for a disproportionate share. Among authors who released additional books with substantial AI text after their first such release, 287 of 385 increased their monthly output after adoption (Fig.~\ref{fig:exposure}c). This expansion was uneven but not confined to a handful of authors: among the 385 identities that released additional books with substantial AI text, the top quartile produced 60.5\% of such post-release books (Gini 0.48). The most prolific authors, however, published at extreme volume, reaching dozens of books a year, and they produced almost entirely books with substantial AI text, with a few exceptions (Fig.~\ref{fig:exposure}d). These authors also reached real commercial scale. The 15 author identities generating the most revenue from books with substantial AI text reached significant commercial volume; books associated with the top identity generated \$1.7M in gross consumer revenue across 8 titles (Fig.~\ref{fig:exposure}e). The 15 highest-revenue individual such books each generated more than \$220K, the top one reaching \$643K on 80{,}431 sales (Fig.~\ref{fig:exposure}f).

Output and revenue are measured for observed byline identity. This means that authors who publish under several pen names are split across these counts, and our approach very likely understates the true amount of production and earnings. Our revenue measures are also calculated before platform commissions, which means the actual author earnings are lower.

\subsection{Successful books with substantial AI text are saturated with distinctive rare language from existing books}

The books that are substantially AI and succeed share a textual signature. To identify it, we measure how much of a book is covered by rare expressions that already occur in existing books, following work that attributes
distinctive language in AI-generated text to large pre-training corpora (the Creativity Index of \cite{lu2025ai} and OLMoTrace of \cite{liu2025olmotrace}). We call an expression rare when it appears in at most five Google Books volumes and is absent from a 4.7-trillion-token snapshot of the internet. The first condition focuses only on language that is present in a handful of books instead of looking into generic and common phrasing, and the second gets rid of expressions that circulate widely online. Therefore, coverage reflects overlap with the distinctive language of particular books \footnote{(Distinctiveness here does not necessarily imply literary quality, however.  Nor does it necessarily mean human-written.)}. This analysis uses a threshold against books that are $>$ 25\% AI, so that each group is clearly distinguishable based on the text. 

 \begin{figure*}[!ht]
    \centering
    \includegraphics[width=\linewidth]{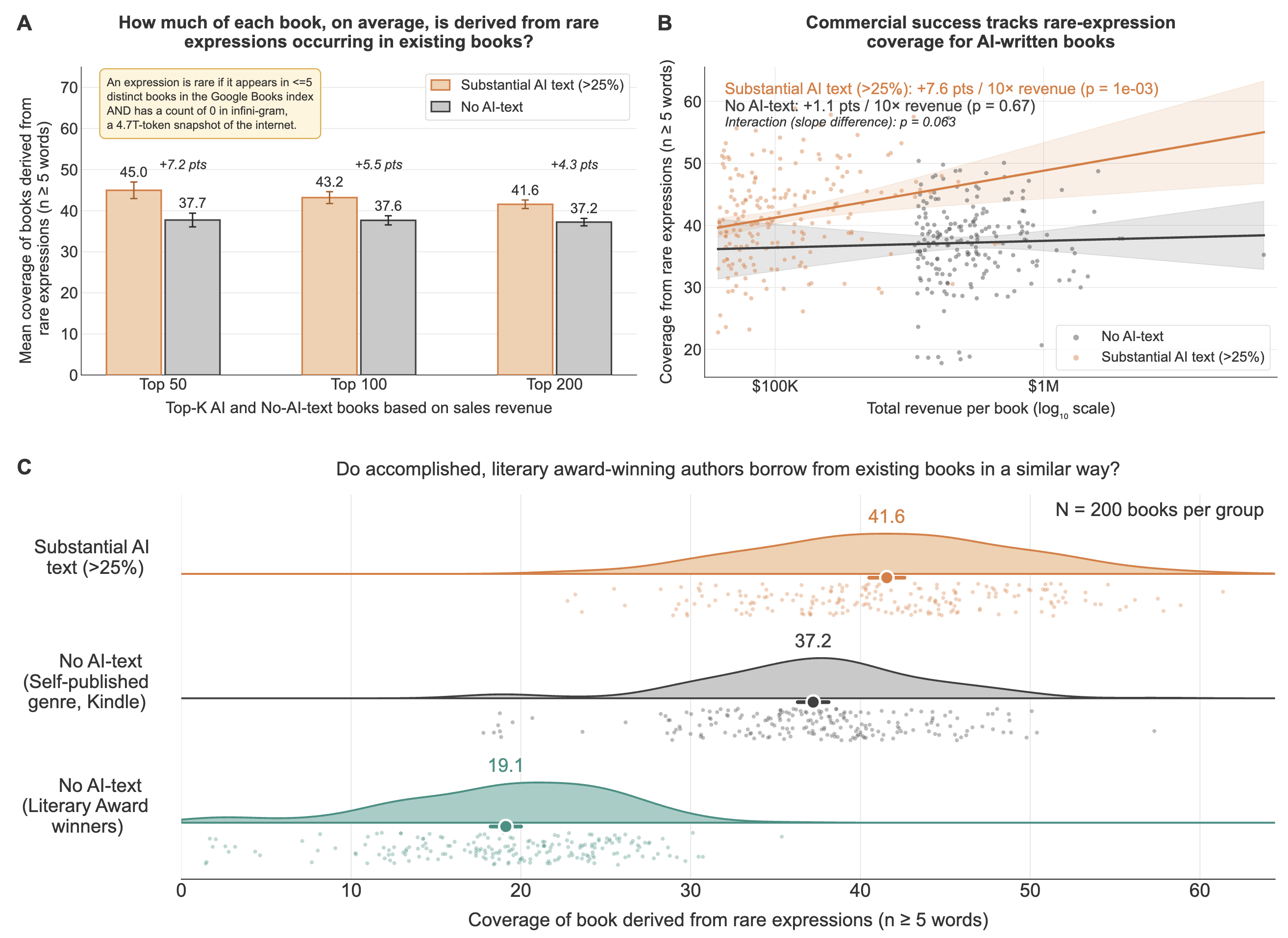}
    \caption{(A) Books with substantial AI text (${>}25\%$) show consistently higher coverage of rare 5+ word expressions than no-AI text books, and the gap widens among the highest earners (+7.2, +5.5, and +4.3 pts for the Top~50/100/200). Books are ranked by total revenue within each category. Error bars = 95\% CI for $n = 50/100/200$ per group; all differences significant (Welch's $t$-test, $p < 10^{-8}$). (B) Among best-selling books with substantial AI text, higher revenue tracks greater reuse of existing books (+7.6 pts per $10\times$ revenue, $p = 10^{-3}$), whereas no such relationship exists for no-AI text bestsellers (+1.1 pts, $p = 0.67$; interaction $p = 0.063$). Points are individual books (200 per group); lines are OLS fits with 95\% CI. (C) Literary award-winning fiction reuses rare expressions far less than either group of bestsellers (means 19.1 vs.\ 37.2 and 41.6). Clouds are kernel density estimates, points are individual books ($n = 200$ per group), and markers show the mean $\pm$ 95\% CI. All pairwise differences are significant (Welch's $t$-test): even the two most similar groups, AI and self-published no-AI text bestsellers, differ at $p = 1.2\times10^{-13}$, and both comparisons against award winners give $p < 10^{-70}$. The individual books being tested are themselves excluded from the Google Books index. Rare expressions are those occurring in $\leq 5$ books in the Google Books index and absent from infini-gram (count~0).}
\label{fig:creativityindex}
\end{figure*}

Considering the universe of top-selling books, the ones that are detected as having substantial AI fill more of their text with these rare expressions than books with no AI text do (Fig.~\ref{fig:creativityindex}a). For the Top-50 tier, the mean incorporation of rare expressions is 45.0\% for books that are substantially AI vs 37.7\% for books with no AI text, that make up for a gap of 7.2 percentage points. The gaps are 5.5 points at Top-100 and 4.3 points at Top-200 \footnote{Welch's t-test, $p < 10^{-8}$}. Comparatively, incorporation of rare expressions for books with no AI text stays flat across tiers, from 37.7\% to 37.2\%. This means that the gap is indicative of the substantially-AI detected books themselves instead of the revenue ranking.

This overlap also correlates with commercial success, but only for books that are substantially AI. For these books, overlap rises with revenue at 7.6 percentage points per tenfold increase ($p = 10^{-3}$), while the slope for books with no AI text is not distinguishable from zero at 1.1 points ($p = 0.67$); the difference between the two slopes is marginal ($p = 0.063$) (Fig.~\ref{fig:creativityindex}b). The steeper slope for the substantially-AI books survives genre fixed effects, so it does not follow from genre composition. 


To test whether drawing specific expressions from existing books is inherent to authorship, we add a third reference group of 200 literary award-winning books across literary, historical, and speculative fiction.\footnote{We consider books that won or were nominated for the Booker Prize, the Pulitzer Prize, the National Book Award, the National Book Critics Circle Award, the PEN/Faulkner Award, and the Nobel Prize.} Award-winning fiction, often held up as a hallmark of originality, has a mean coverage of just 19.1\%, roughly half the rate for books that are substantially AI (41.6\%) and well below self-published bestsellers with no AI text (37.2\%); all pairwise differences are statistically significant
(Fig.~\ref{fig:creativityindex}c).

The rare expressions derived from existing books are surprisingly embodied in books with substantial AI text. Examples of rare expressions found in AI books are ``cheekbones more pronounced, the shadows beneath her eyes deeper", ``The question circled his mind like a vulture", ``cross my eyes and puff out my cheeks like a demented chipmunk" and ``my stomach flip-flopping like a fish out of water". These expressions route meaning through touch, smell, and visceral sensation. Our findings align with recent work from \cite{russell2026storyscope} showing how AI overwrites the body and senses, conveying feeling through physical sensation and environment while rarely labeling emotion explicitly.

\section{Discussion}
Research on generative AI in creative work has focused primarily on whether machine output is good enough to substitute for human work \cite{chakrabarty2025readers,chakrabarty2026can}. Our results, however, point to a different problem for creative markets, one that is agnostic to the quality of AI work. Our analysis of more than 14,000 self-published genre-fiction books sold on Amazon between 2023-26 shows that the catalog grew far faster than the sales and revenue. That said, total revenue also rose over this period, but at a much slower pace, and as a result the revenue per book declined. Books containing AI text were largely responsible for this substantial increase in the growth, but most such books sold very few copies. We further observe that books with no AI text  earned less per book in later cohorts as well, so our results cannot be explained by mix-shift alone~\cite{foster2001aggregate,baily1992productivity}, where the average falls only because the catalog is filled with many low-earning AI books. Per-book revenue fell most in the genres where AI text had diffused the most. These patterns are consistent with dilution as each book earns less after substitutable books crowd a market. Hence, generative AI reshaped the book market without competing on quality at all, displacing revenue and top ranks through sheer volume of books alone.

Two factors shaped this outcome, one on each side of the market. On the supply side, AI adoption increased output among authors who continued using it. Many authors who adopted AI did not publish more, but among those that released further books containing substantial AI text, most increased their output, and a smaller set of them produced dozens of books a year. A prolific minority achieved real commercial success, with books associated with some individual producers generating more than a million dollars in gross consumer revenue. On the demand side, a smaller subset of books containing substantial AI text attracted substantial consumer demand and reached the top of the market. Among these high-revenue books, rare-expression overlap with existing books was greater than among top-selling books with no detected AI text, and, within the substantial-AI group, overlap rose with revenue. This pattern is consistent with a derivative-production channel\cite{landes1989economic, landes2003economic}, though the measure captures textual overlap and not copying of any particular book \footnote{Successful AI books appear to draw commercial value from recombining distinctive expression found in existing works, though our measure shows only aggregate overlap with prior books and cannot attribute any passage to or establish copying of a specific title}. These two forces compounded each other: cheap production lengthened the tail of supply, and the rare-expression overlap characterized the smaller set of AI books that reached the top.

Our results extend recent work on AI and book markets. Reimers and Waldfogel \cite{reimers2026ai} showed that books flagged as AI from short previews earn lower ratings and weaker sales than non-AI books. In contrast, we detect AI across the full text of each book and document how the entry of these books affects the outcomes for other books in the marketplace. This difference is important because we find that the books containing AI text sell relatively little on average, but since they are so numerous, they have the ability to move the market. On the supply side, authors are affected as they compete for sales and visibility in this crowded marketplace. However, on the demand side, end consumers might benefit from the greater variety and lower prices of books, so the net welfare effect is unclear. But for the questions we address, i.e., the returns to human creative effort, and the market-dilution issue at the heart of copyright law, the harm to authors is what is at stake.

These findings bear on questions now before the courts. In recent litigation over training large language models on copyrighted books, market effect has emerged as the central unresolved empirical question.  The “market dilution” theory endorsed by the Copyright Office \cite{copyright2024ai} and entertained by Judge Chhabria in Kadrey v. Meta~\cite{kadrey2025meta},  recognizes that human-authored works may be ``crowded out'' by AI-generated outputs which may substitute for sales of the human-authored works on which the AI platform trained (i.e., human-authored works copied into the training data).  Liability (if any) for infringement results not from the platform’s production of outputs that compete with human-authored works, but from the massive upstream copying that enabled the competing outputs.\footnote{The outputs may also compete with human-written works that the AI system did not copy, but that substitution effect is not within the purview of copyright law: competition without copying is not infringement} In Kadrey v. Meta, the plaintiff authors failed to allege facts supporting a substitution effect from market-flooding by AI titles.

In Kadrey v. Meta, concerning proof of market dilution, Judge Chhabria inquired whether AI-generated and human-written books compete in the same markets, and ``what impact does this competition actually have on sales of the books it competes with? Whatever the effects have been thus far, are they likely to increase in the future, as more and more AI-generated books are written, and as LLMs get better and better at writing human-like text?''  This study examines the market for self-published genre titles on Amazon, a market in which human-written books and books with substantial AI content compete. It documents the depressive effect of the massive entry of AI content titles on the sales of human-authored books.  Because the effect is one of scale, one may surmise that ``as more and more AI-generated books are written,'' the crowding effect will increase.

We emphasize that the study ties the diminution of human-authored books' earnings to the volume of market entry of AI titles, without regard to their comparative quality. The study rules out the conclusion that the average revenue per book fell merely because the catalog filled up with a plethora of low-earning AI books, cumulatively depressing the average without deleteriously affecting the sales of human-written books. In other words, bloating the catalog with poorly selling AI books is not akin to the effect on individual students in a college class when weaker students enroll: there, the overall grades fall, but the weaker students' results do not drag down the stronger students' grades. Rather, this study shows that books with ``no AI text" (the good students) also, as a category, earned less than human-written books had earned before AI-content books began to compete with them. AI books' entry into the market on a vast scale does not just overpopulate the bottom tier of sales; it allows at least some of the entrants to vie for the higher echelons of revenue generation, to the displacement of human-authored content. We note that this study does not attribute a reason for competition through scale; it merely documents the phenomenon. One might speculate that the absence of disclosure of significant AI content may give the new entrants a competitive advantage, or allow them to elude a competitive disadvantage, assuming that readers informed of the presence of AI content will be less inclined to buy the book. We do not, however, have the data that would permit us to draw that conclusion. It is also possible that the pen names of some producers of AI books acquire a kind of brand name recognition: if AI-aided “Alice Author” has a following, the surfeit of books generated under that pseudonym, relative to human authors, may better satisfy audience demand for more titles with less waiting for new titles to appear, and may thus increase sales attributable both to volume and to brand loyalty.  This outcome may resemble results obtained by brand-name authors who employ a stable of human ghostwriters to keep up with reader demand, but, as with AI book production generally, the scale and speed of production far outpace the endeavors of even multiple humans.  Here again, however, we lack data addressing the brand-name effect of particularly prolific AI pseudonyms.

Our comparisons are associational. We compare release cohorts and relate outcomes to genre-level AI exposure, and genres differ along many dimensions, so we read these comparisons as observational. By dilution we mean two declines we observe as the catalog fills with AI books, in per-book revenue and in the market share of books with no detected AI text. Our headline finding, the fall in revenue per book, is present even within books that have no detected AI text, so it cannot be explained by the entry of low-earning AI books alone. On its own this might seem consistent with a broad market trend, but the pattern reverses in Fantasy / Supernatural / Horror, the genre AI reached latest and least, which makes it unlikely to be caused by a market-wide trend. Our byline-level measure of output and revenue understates concentration when one author writes under several names, so the concentration we report is a lower bound. And the magnitudes we report are specific to the single platform we observe, though their direction is not.

Our findings extend well beyond self-published fiction. The conditions that made this market vulnerable, i.e., low entry costs, readers returning for familiar kinds of books and authors, discovering new books via a shared pool of rankings and recommendations where books with and without AI text compete directly without disclosure, are not unique to it.  We should observe similar outcomes to those in our study in other domains where these conditions hold, i.e., a near-limitless supply of cheap works meets a fixed reserve of reader attention. Music, stock media~\cite{goldberg2025generative,dai2026generative}, and short-form writing markets share several of these features.  Our work provides the first systematic evidence of these effects in self-published genre fiction, and our results have direct implications for choices now facing authors, platforms, and courts.

\section{Methods}

\subsection{Observed market panel} 

We study sales of newly self-published genre-fiction e-books released between January 1, 2023, and March 31, 2026.\footnote{During publisher verification we identified a small number of traditionally published editions that accounts for roughly 0.3\% of the corpus. We retain them in the analysis.} We have a proprietary sales panel that includes daily sales as well as Amazon sales rankings until June 29, 2026, which is 90 days after the last release date in our window, so that even titles released at the end of the period can be tracked through their full launch window. The sales panel was provided by one of the ``Big Five'' publishers. We chose to start with the entire population of self-published works (newly released between January 2023 and March 2026) and then reduced the search space by excluding titles whose Amazon sales rank never fell below 1.5 million at any point during the observation period. This got rid of listings that basically have no sales activity. We drew a stratified random sample of 15,000 titles from the remaining pool of eligible titles based on their genre classification. This number was the maximum feasible amount given the computational cost of running AI detection on full texts. We divided the samples across different genres w.r.t to the rough proportion of their contribution to the overall number of new releases (General Romance 28.8\%, Speculative / Adventure Romance 15.5\%, Fantasy / Supernatural / Horror 15.2\%, General / Contemporary Fiction 11.4\%, Mystery / Thriller / Crime 10.7\%, Crime / Suspense Romance 9.6\%, Sports Romance 5.8\%, Science Fiction / Adventure / Dystopian 3.1\%), so that the sample reflects the composition of the release population. We also stratified the distribution of titles according to when they were released. We selected titles blind to content; thus, the results of our detection process could not affect whether a title is included. Finally, some of our selected titles had no usable sales data recorded in the panel and were dropped. Thus, we ultimately used 14,419 titles in our analyses.

In general, throughout this paper we refer to the ``title'' as the ASIN level catalog entry that we are analyzing and refer to the ``book'' as the actual work being read. Each title has its unique Amazon Standard Identification Number (ASIN), along with author and title metadata, an on-sale date, subject and genre categories, Kindle Unlimited fields, daily Amazon unit sales, consumer price, Amazon sales rank, and a score giving the share of its full text detected as AI-generated. Additional details regarding these fields, and how we construct the main variables, can be found in the Supplementary Information.

Our panel consists of only those titles that appear in Amazon's sales and ranking feeds, thereby limiting our observations to only those titles that actively participate in Amazon's marketplace. All the estimates reported in the paper, such as the catalog shares or the dilution estimates are based on this observed universe of books only.  Next, we group all titles into eight main genre clusters and assign the remainder to a miscellaneous “Other” category that we exclude from the main figures. Details are provided in the Supplementary Information. Additionally, since we want to compare the performance of titles over equivalent time periods (i.e., launch windows), those comparisons are restricted to the titles and release cohorts observed for the full length of their respective launch windows.

\subsection{Panel construction and revenue}

We parse the date, numeric, and score columns from the raw data into their respective types, trim the title ID strings to remove excess whitespace, and remove empty IDs. We then create a pair of connected tables, one at the title level and one at the title-day level. The first table is a list of all titles with each title having one row in this table. For any given title, if there are multiple rows that contain different information (i.e., more than one possible value) for any of the title metadata we retain the most frequent non-missing value for each field. The second table is a list of all days by title. In this case each row represents an observation of a particular title on a specific day. Any duplicate rows of this nature are collapsed down to just one observation for each day.

To calculate daily revenue for title $i$ on day $t$ we use:
\begin{equation*}
R_{it} = Q_{it} \times P_{it},
\end{equation*}
where
\begin{itemize}
\item $Q_{it}$ equals the quantity of units sold.
\item $P_{it}$ equals the consumer price for those units.
\end{itemize}
We replace any $P_{it}$ that are missing or non-positive with the median valid price within the time period of the title's launch window. All titles have complete unit sales and price data during the entire launch window (the observed preorder days before the on-sale date plus days 0 through 90 after release).

\subsection{Full-text AI classification}

We use Pangram, one of the most accurate AI detectors audited by several independent research groups \cite{jabarian2025artificial,russell2025people,shah2026access} and now used by the popular online publishing platform Substack. Each book carries a full-text AI score from the Pangram detector. We segment each book into individual chapters after excluding Contents and Acknowledgements, Dedications, Copyright pages, Author bios, Previews and pass each chapter to Pangram's v3.3 API. Pangram's detection algorithm returns a list of analyzed text windows spanning the input chapter. Each window includes text, label, ai\_assistance\_score, confidence, start\_index, end\_index, word\_count, and token\_length. Across all chapters of a book, we count the windows whose label is not ``Human Written'' and define the AI-detection score as the percentage of windows so flagged,
\[
P_i \;=\; 100 \times
\frac{\left| \{\, w \in W_i : \ell(w) \neq \text{``Human Written''} \,\} \right|}
     {|W_i|},
\]

where $W_i$ is the set of analyzed windows for book $i$ and $\ell(w)$ the label of window $w$. We assign each title to one of three bands based on $P_i$,
\[
\text{No AI text}_i = \mathbf{1}[P_i = 0],
\quad
\text{Light AI text}_i = \mathbf{1}[0 < P_i \leq 25],
\quad
\text{Substantial AI text}_i = \mathbf{1}[P_i > 25].
\]

A book with no AI text has no flagged windows; one with light AI text has a flagged share above 0\% and at most 25\%; one with substantial AI text has a flagged share greater than 25\%. These bands are discretized versions of continuous scores $P_i$. For our study, we used Pangram version 3.3~\cite{pangram33} which was released in May 2026. This latest version shows improved performance (recall) on long-form  (2,000+ words) text, humanized text, and has a smaller false positive rate (FPR) on non-native English writing. Its performance has been validated on a sample containing 566,745 texts drawn from 15 domains. Held-out evaluation reports a $N$-weighted false-positive rate of 0.04\% and a false-negative rate of 0.14\%.

\subsection{Genre, platform, and author variables}

Amazon provides a detailed list of genres and subjects for books published on its platform. However, its genre classification is too fine-grained for performing analyses at the level of genre-month since many cells will have too few books. So, we clustered the raw Amazon genre labels into eight broad groups: General Romance, Speculative/Adventure Romance, Crime/Suspense Romance, Sports Romance, Fantasy/Supernatural/Horror, Mystery/Thriller/Crime, Science Fiction/Adventure/Dystopian, and General/Contemporary Fiction. These genre groupings were developed to retain as much commercially relevant signal as possible and they ensured that each genre-month cell contains enough books to support sufficiently powered analyses. These genre groupings were fixed prior to performing any empirical work. The Supplementary Material contains the conversion of raw Amazon genre labels to the eight groups defined above.

Next, for each book title in our sample we defined the Kindle Unlimited (KU) flag based on whether the title ever participated in the program. So, when performing the genre-month analyses we computed the KU-share as the fraction of titles released in that genre in that month that had the KU flag checked. Unfortunately, our panel data does not break down the number of unit sales of a book into KU page reads and à la carte purchases.  

Finally, we constructed author identity using information contained in the byline field of each book title. So, books with missing bylines were not part of our author-level analysis. We anonymized the authors in the figures in the paper as A1, A2, and so on. It is worth noting that one potential shortcoming of our setup is that an author may write under several pen names and several people may share one byline. Hence, the byline count may underestimate how concentrated a given author's production is.

\subsection{Market outcomes and dilution}

How do the title shares of each group translate into sales and revenue? We use three measures for that -- title share, sales share, and revenue share of each band. For band $h$
\begin{align*}
\text{title-share}_h &= N_h/N\\
\text{sales-share}_h &= \left(\sum_{i\in h}\sum_t Q_{it}\right)\Big/\left(\sum_i\sum_t Q_{it}\right)\\
\text{revenue-share}_h &= \left(\sum_{i\in h}\sum_t R_{it}\right)\Big/\left(\sum_i\sum_t R_{it}\right)
\end{align*}

Much of what follows uses a fixed window around the time of the title's launch, defined as the observed preorder days before the title goes on sale along with days 0 through 90 after it,
\[
\mathcal{L}_i = \{t : t < O_i\} \cup \{t : 0 \leq t - O_i \leq 90\},
\]
where $O_i$ is the title's on-sale date. The fixed 90-day horizon gives every title the same follow-up period after its release. We include preorder days because Amazon records sales and ranks for some titles before their on-sale date. Within each title's launch window we compute its launch sales, launch revenue, days with positive sales, and best rank. For our analysis pertaining to the books in the upper tail, i.e., top 50\%, 25\%, 10\%, 5\%, and 1\% we sort the titles by launch sales.

We measure dilution in two ways. The first measure tracks the market on a quarterly basis. For each quarter $q$ we count released titles, titles with positive sales, total sales, and total revenue, and index each series to its 2023 Q1 level. We also compute revenue per selling title, i.e., titles that sold at least one copy,
\[
\text{RevPerSellingTitle}_q = \frac{\sum_i \sum_{t \in q} R_{it}}{\#\{i : \sum_{t \in q} Q_{it} > 0\}}.
\]
Because every title on sale in a given quarter enters this measure, it represents a snapshot of the market in which older and newer titles compete simultaneously. The second measure compares each title over a common time frame. We compute the launch window revenue for each book that sold at least one copy over that time period, then, group titles by release quarter, and retain only the release cohorts with complete launch-window data. So, our measure only accounts for a release quarter when its last-released title can be observed through day 90 after its on-sale date. Finally, we report this metric for the entire marketplace, broken down by AI text band and genre, and for books without any AI content.

\subsection{Rank-slot and exposure analyses}

We use the Amazon sales rank as an indicator of market positioning, where a lower number indicates a better position. We cannot see the internal signals that influence impressions or placements of books on the platform, such as recommendation algorithms, search result rankings, and paid advertising, so our rank-slot analyses relate solely to observed sales-rank positions.

Within each genre-week cell, we take each book's median observed sales rank, order titles by that median, and define Top-10 and Top-25 rank slots.
For the genre heatmap, we compute the share of Top-25 slots held by books with substantial AI text in each genre-quarter cell.

Next, for the top-rank churn analysis (Figure 2D), we compare the sets of Top-25 book titles from one quarter to the next. Let $T_{gq}^{25}$ denote the titles holding Top-25 rank slots in genre $g$ during quarter $q$. Retention of books with no AI text is the fraction of the prior quarter's no-AI Top-25 titles that remain in the current quarter's Top-25, 
\[
\text{Retention}_{gq}
=
\frac{
\left| \{i \in T_{g,q-1}^{25} : i \text{ has no AI text}\} \cap T_{gq}^{25} \right|
}{
\left| \{i \in T_{g,q-1}^{25} : i \text{ has no AI text}\} \right|
},
\]
and the entrant share is the fraction of new Top-25 entrants with substantial AI text,
\[
\text{EntrantShare}_{gq}
=
\frac{
\left| \{i \in T_{gq}^{25} \setminus T_{g,q-1}^{25} : i \text{ has substantial AI text}\} \right|
}{
\left| T_{gq}^{25} \setminus T_{g,q-1}^{25} \right|
}.
\]
Because the first quarter does not have any preceding quarter to compare against, we exclude it from our analyses of churn.

To perform AI-exposure analyses described in Figures 4A,B, we first construct a genre-month level panel. Then, for every cell (genre-month) in that panel we calculate, a) the share of the catalog containing books with substantial AI text; b) the share of sales of books with no AI text; c) the share of Top-25 slots occupied by books with no AI text; d) the Kindle Unlimited (KU) share of books; and e) several covariates listed in detail in the Supplementary Information. Next, in Figure 4A, we group all the genre-month cells into bins based on the share of books with substantial AI in each of them and plot the average Top-25 share held by books with no AI text in each bin. The 95\% confidence intervals are calculated as $\bar{y} \pm 1.96\,\text{SE}$.

Then, in Figure 4B, we test how the no-AI lead changes with Kindle Unlimited availability. We rank the genre clusters by their KU share and divide them into a KU-light group and a KU-heavy group. We drop the middle. Within each KU group, we define the lead as the difference in sales or revenue share between books with no AI text and books with substantial AI text. We then estimate how it differs between the two groups by controlling for AI exposure and calendar month. The main figure reports the adjusted shares of each band in each KU group, and the difference in the no-AI lead between the groups, for both sales and revenue. We compute confidence intervals by resampling genre clusters. We read this difference as heterogeneity by KU availability across genres, and not as a causal effect of Kindle Unlimited.

\subsection{Producer-scaling analysis}

We use the byline identifier developed above to track author output around the time-period that they adopted AI. For each author, their adoption month is computed as the first month in which they publish a book containing substantial AI text. Then, for every author who produced at least one such book during the study period, we build an author-month panel containing the titles they released in each month, separately accounting for the titles with substantial AI text and other titles.

We report the number of titles that each author produced in event time, relative to each author's AI adoption month. The pre-adoption period covers the time prior to when the author adopted AI and the post-adoption period covers months 1-12 following their adoption. We ignore the adoption month, because it can conflate the first title with substantial AI text with other titles, potentially with no AI text, from the same production burst. We also exclude authors whose panels do not have at least 3 months in the post-adoption period. For each remaining author, we calculate the output per month before and after adoption, the titles with substantial AI text and other titles published in the post-adoption period, and the change in monthly output across the two periods.

Further, we define observed existing authors as those authors who had at least one observed title prior to their adoption month, and newly observed authors as those who did not. This second category describes only our study panel and does not indicate that the author published nothing elsewhere.

We rank all authors by the increase in their average monthly volume of titles published after adoption and display the 15 greatest gains, breaking down their total post-adoption output into titles with substantial AI text and other titles. We assign each of these authors a single dominant post-adoption genre, the genre in which they publish the most titles after adoption.

We rank author identities by the revenue their books with substantial AI text generate during the 12 months following adoption, and we rank individual books with substantial AI text by the revenue each earns in its own first 12 months on sale. In both these analyses we keep only those authors and books that had a complete 12-month follow-up, and we display the top 15 of each. Note that both these analyses report a 12-month horizon (the author ranking follows each author's adoption clock and the book ranking follows each book's own release date). This 12-month horizon differs from the 90-day launch window used in the market-level analyses. Concentration metrics for these two rankings are reported in the Supplementary Information; specifically, we report both the Gini coefficient and the inverse-Herfindahl effective count.

\subsection{Rare-expression overlap}

We measure rare-expression overlap using the full text of the top-revenue books with substantial AI text ($>25\%$) and those with no AI text. Each book was broken down into word sequences to enumerate all expressions containing at least five words. A particular expression is considered to be an example of rare existing-book language if both of the below criteria were met: it appeared in no more than five different volumes within the Google Books index; it did not appear anywhere within infini-gram, a 4.7-trillion-token web snapshot. The first criterion identifies language used in a very limited set of publications; the second eliminates expressions which occur frequently in general web-based language.

Following this, we calculate rare-expression coverage for each book as the ratio of the number of tokens flagged as being part of at least one rare expression to the total number of tokens within that volume \cite{liu2026alignment}. Coverage is measured at the token level: every word within a given book is either covered by at least one rare expression, or not. Because we flag words rather than counting instances of expressions, overlapping matches do not inflate the total, and coverage remains bounded between 0 and 100\%.

To see if such overlap corresponds to commercial success, we fit a linear model to the top 200 books from each group:
\[
C_i =
\alpha +
\beta_1 AI_i +
\beta_2 \log_{10}(Revenue_i) +
\beta_3 AI_i \times \log_{10}(Revenue_i)
+ \epsilon_i.
\]
Here, $C_i$ is rare-expression coverage and $AI_i$ is a binary variable marking books with substantial AI text (i.e., ${>}25\%$). The parameter $\beta_2$ represents the revenue slope for books with no AI text and $\beta_2 + \beta_3$ the slope for books with substantial AI text, so $\beta_3$ measures how much more steeply coverage rises with revenue for AI-heavy books than for books with no AI text.

The Supplementary Information provides further detail regarding our tokenization and query methods, how we handled potentially overlapping expressions and near-duplicate or series titles, as well as the complete regression results.

\subsection{Statistical analysis}

We calculate all major market variables from observed total values. These include band shares of titles, sales, and revenue, top-rank slot shares, calendar-quarter growth multiples, and launch-window revenue per selling title. Calendar-quarter growth multiples index each series to its Q1 2023 level. Launch-window revenue per selling title is defined as total launch-window revenue divided by the number of titles with positive observed unit sales in that window. To create the binned exposure plots, we provide bin averages and corresponding 95\% interval estimates based on the formula $\bar{y} \pm 1.96\,\text{SE}$.

For the Kindle Unlimited and the rare-expression analyses, we employ linear regression models. The Kindle Unlimited model uses genre-month observations with month and exposure-bin fixed effects; we estimate the difference in the no-AI text lead over books with substantial AI text between KU-heavy and KU-light genres, for both sales and revenue share, using confidence intervals from a genre-cluster bootstrap. For the rare-expression analysis we compare Top-$K$ means with Welch's $t$-tests and estimate the interaction between substantial-AI status and $\log_{10}$ revenue using ordinary least squares (OLS), adding genre fixed effects to our controlled model; we also provide effect sizes in percentage points of book coverage. This analysis compares books with no AI text against books that are more than 25\% AI. Results for varying thresholds relative to this 25\% market cut-off are reported in the section \ref{cutoff} as well as in Supplementary Information, along with regression results for both the Kindle Unlimited and rare-expression analyses.

\subsection{Does the 25\% cutoff matter? \label{cutoff}}

We call a book substantial AI text when more than 25\% of its text is flagged by the detector. This leaves us with an open question on how much of our result depends on this threshold. Table \ref{tab:s4_threshold_sweep} provides detail on  varying the detectability thresholds and reports the share of titles, sales, and revenue that fall in the substantial band at each setting. The picture holds throughout. This means that while a higher cutoff puts fewer books in the substantial AI band, at every setting the band takes a smaller share of sales and revenue than of titles.

\begin{table}[!htbp]
\centering
\caption{The substantial band at different cutoffs, over the launch window.}
\label{tab:s4_threshold_sweep}
\begin{tabular}{lrrrr}
\toprule
Cutoff & Titles & Title share (\%) & Sales share (\%) & Revenue share (\%) \\
\midrule
$S>10\%$ & 3,586 & 24.9 & 16.1 & 15.2 \\
$S>20\%$ & 3,066 & 21.3 & 13.1 & 12.2 \\
$S>25\%$ & 2,880 & 20.0 & 12.1 & 11.3 \\
$S>30\%$ & 2,741 & 19.0 & 11.0 & 10.2 \\
$S>40\%$ & 2,440 & 16.9 & 9.4 & 8.6 \\
$S>50\%$ & 2,168 & 15.0 & 7.9 & 7.1 \\
\bottomrule
\end{tabular}
\end{table}

\section*{AI Disclosures} In accordance with responsible AI-disclosure authors of the paper acknowledge that some parts of the papers were copy-edited by AI. Authors of the paper verified every detail for consistency and correctness.

\section*{Data availability}
The full texts analyzed in this study are copyrighted books and cannot be shared publicly; they are available from the corresponding authors upon reasonable request. The daily sales panel is proprietary and was provided under a data use agreement that prohibits public release. The aggregate data underlying all figures is available from the corresponding authors upon reasonable request.

\section*{Code availability}
All code used for data processing, classification, and statistical analysis is available at \url{https://github.com/tuhinjubcse/BookSlop}.

\section*{Supplementary Information}

Supplementary information for the paper can be found on this link \url{https://tuhinjubcse.github.io/Supplementary.pdf}

\section*{Ethics statement}
All the books analyzed in our study were lawfully acquired through various sources such as emailing authors for epubs, acquiring online through libraries such as Libby or Hoopla or purchasing directly from Amazon. We report author-level analyses using anonymized identifiers (A1, A2, \ldots) and do not name any author, byline, or book title. The study analyzes market-level outcomes from observational sales data and involved no interaction with human subjects.

\bibliography{science_template} 
\bibliographystyle{naturemag}




\clearpage

\end{document}